\documentclass{article}

\usepackage{arxiv}

\usepackage[utf8]{inputenc} 
\usepackage[T1]{fontenc}    
\usepackage{hyperref}       
\usepackage{url}            
\usepackage{booktabs}       
\usepackage{amsfonts}       
\usepackage{nicefrac}       
\usepackage{microtype}      
\usepackage{lipsum}
\usepackage{graphicx}
\usepackage{subfigure}
\usepackage{booktabs}
\usepackage{amsmath}
\usepackage{amssymb}
\usepackage{mathtools}
\usepackage{amsthm}
\usepackage{diagbox}
\usepackage{authblk}
\usepackage[capitalize,noabbrev]{cleveref}
\graphicspath{ {./images/} }

\title{Coordinate-Aligned Multi-Camera Collaboration \\ for Active Multi-Object Tracking}

\author[1]{Zeyu Fang$^*$}
\author[1]{Jian Zhao$^*$}
\author[1]{Mingyu Yang}
\author[1]{Wengang Zhou}
\author[2]{Zhenbo Lu}
\author[1]{Houqiang Li}

\affil[1]{University of Science and Technology of China}
\affil[2]{Institution of Aritificial Intelligence, Hefei Comprehensive National Science Center}

\begin{document}
\maketitle
\def\thefootnote{*}\footnotetext{These authors contributed equally to this work}
\begin{abstract}
Active Multi-Object Tracking (AMOT) is a task where cameras are controlled by a centralized system to adjust their poses automatically and collaboratively so as to maximize the coverage of targets in their shared visual field. 
In AMOT, each camera only receives partial information from its observation, which may mislead cameras to take locally optimal action.
Besides, the global goal, \emph{i.e.,} maximum coverage of objects, is hard to be directly optimized.
To address the above issues, we propose a coordinate-aligned multi-camera collaboration system for AMOT.
In our approach, we regard each camera as an agent and address AMOT with a multi-agent reinforcement learning solution. 
To represent the observation of each agent, we first  identify the targets in the camera view with an image detector, and then align the coordinates of the targets in 3D environment. 
We define the reward of each agent based on both global coverage as well as four individual reward terms. 
The action policy of the agents is derived with a value-based Q-network. 
To the best of our knowledge, we are the first to study the AMOT task.
To train and evaluate the efficacy of our system, we build a virtual yet credible 3D environment, named ``Soccer Court'', to mimic the real-world AMOT scenario.
The experimental results show that our system achieves a coverage of 71.88\%, outperforming the baseline method by 8.9\%.
\end{abstract}


\section{Introduction}
With recent progress in computer vision and robotics, Active Object Tracking (AOT) has become an emerging  task~\cite{luo2018end,zhong2018ad,li2020pose}, in which mobile cameras track objects effectively by adjusting their poses based on the visual observations automatically.
Existing works in the field of AOT only focus on single-object tracking task.
However, there are various real-world applications requiring multiple cameras to track multiple objects, such as sports competitions, traffic monitoring, \emph{etc}.
Considering the potential application value, in this work, we are dedicated to Active Multi-Object Tracking (AMOT) task, where a team of collaborative cameras automatically control their actions to track multiple objects, and approach it with a multi-camera collaboration system.

Compared to the settings of existing works where each camera is responsible for tracking a single object, AMOT requires multiple cameras to coordinate the attention to multiple objects.
AMOT is a challenging task due to the following facts. 
Firstly, AMOT involves multiple cameras to track the targets, and the problem complexity grows exponentially with the camera number. 
Secondly, each camera only captures a local and partial observation of the scene filed, which requires an effective information integration scheme to achieve global optimization of camera coordination. 
Thirdly, the evaluation metric, \emph{i.e.,} overall target coverage, is somewhat high-level, which is difficult to evaluate the control action on each individual camera and guide the cameras to learn an effective collaboration. 

\begin{figure}
    \centering
    \includegraphics[width=\columnwidth]{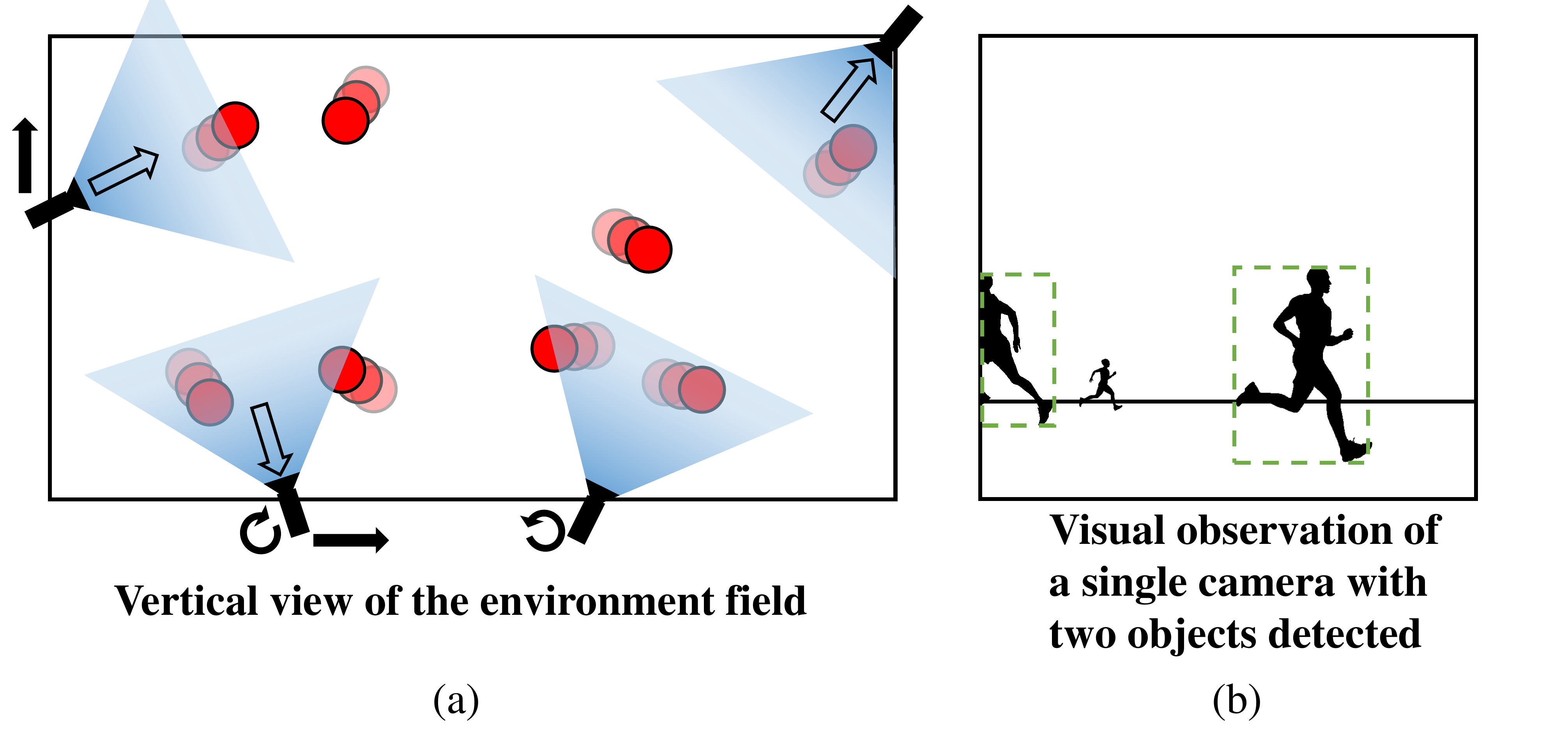}
    \caption{An overview of our environment for AMOT task. (a) A vertical view of the global situation. Multiple randomly moving targets are tracked by a team of cameras. Based on its visual observation and policy, cameras take actions including translation, rotation and zoom to cover maximum number of targets in global visual field. (b) A typical view observation of a single camera. Bounding boxes of each target are generated by object detection algorithm. A target is considered to be covered only if the size of its bounding box is larger than a threshold.}
    \label{fig:introduction}
\end{figure}

To address the above issues, we formulate AMOT as a centralized multi-agent reinforcement learning (MARL) problem and propose a coordinate-aligned multi-camera collaboration system.
Specifically, we regard each camera as an agent and subtly define the state representation and reward function for the agents. For the state representation, considering that agents (cameras) directly capture video shots of the scene field from different perspectives, we first identify the targets in the video frames with an image object detector~\cite{wang2021scaledyolov4}, and then align the detected targets by camera calibration \cite{786974}. In this way, the targets captured by all agents are mapped to the same absolute coordinate system in the 3D environment, which integrates the observation from all agents into a global state representation. To effectively train the agents with reinforcement learning, we design the reward function with both global and local terms. The global reward term is defined as the coverage rate of all targets, while the local reward terms are specifically defined for each agent considering the visibility, direction, bounding box and position rewards of the corresponding targets in the view field. 



Based on the above definition, we leverage a value-based reinforcement learning algorithm to identify the optimal policy so as to select a proper action to control each agent (camera). 
Specifically, a shared deep Q-network \cite{van2016deep} is used to estimate the Q-value function, which evaluates the Q-value of each possible camera action given integrated state information including coordinates as observations.
In this way, agents learn to collaborate effectively, and meanwhile, the individual-global gap is largely narrowed.


To train and evaluate our proposed system, we build a 3D environment for the AMOT task, as illustrated in Figure~\ref{fig:introduction}, based on Unreal Engine~\cite{qiu2017unrealcv}. 
Different from the existing 2D environment \cite{xu2020learning}, which lacks visual appearance and complex scenes like occlusion, our environment is virtual yet credible with many human targets to mimic the real-world scenarios.
Our environment settings are defined according to real-world soccer match discipline, where multiple players are distributed in a large court walking during a whole episode. Multiple cameras are evenly placed at the border and controlled to cover as many targets as possible.
Such an environment enables our system to acquire visual observations and conduct actions on cameras, and also provides precise rewards, facilitating the training.
The environment can efficiently render the next state (\emph{i.e.,} the next frames) based on the current environmental situation and conducted actions.

We validate the effectiveness of our method by conducting experiments in the environment mentioned above.
The results show that our method achieves considerable performance gain compared to the baseline method in which cameras are fixed.
Moreover, the ablation study demonstrates the effectiveness of the inverse projection transformation design and the necessity of each type of individual reward.

Our contributions are summarized as follows:
\begin{itemize}
  \item [1)] 
  To the best of our knowledge, we are the first to study the AMOT task and propose a solution based on centralized multi-agent reinforcement learning, which learns an efficient collaborative tracking policy for all the involved cameras. This solution will serve as an initial baseline in this field. 
  \item [2)]
  We build a 3D virtual environment to mimic real-world multi-object active tracking scenes, which enables the training of agents on active tracking tasks and also has a potential of generalization to more complicated scenes. Our virtual environment is released to the public for further study in the community. 
\end{itemize}

\section{Related Work}
In this section, we discuss related work from three aspects: traditional solutions to AOT, and deep RL for single-camera and multi-camera AOT. 

\subsubsection{Traditional solutions to AOT} Traditional solutions to active tracking usually consist of two separate modules, \emph{i.e.,} detection and control.
The detection module first detects and analyzes motion and location features, which are then used by the control module to manipulate the camera sequentially \cite{active1462901,denzler1994active,856853,murray1994motion}. 
Thanks to the great progress achieved in conventional object detection and tracking in recent decades \cite{li2019robust,bochkovskiy2020yolov4,8419331,8457310,8283783}, passive tracking tasks, in which the video data is pre-captured, can be effectively addressed with existing tracking algorithms.
According to these tracking results, a control module is developed to manipulate the camera and track targets actively \cite{8090332,4653062}.
Similarly in our method, an object detector is also used to obtain visual observations as the input of RL algorithm.
However, traditional methods suffer limited generalization capability due to the lack of data sets.

\subsubsection{Single-camera AOT via deep RL} Recently the breakthrough in deep reinforcement learning \cite{mnih2013DQN,mnih2015human,hessel2018rainbow} and multi-agent reinforcement learning \cite{rashid2018qmix,zhang2021multi} provides an alternative way to deal with the active tracking problem. 
Previous works show that the RL algorithm is feasible and data-efficient for many vision-based control tasks \cite{wu2016training,pan2017virtual,zhu2017target,hong2018virtual}, which inspires researchers to employ RL algorithm on visual tracking problem.
In active tracking task, the RL algorithm is first applied via an end-to-end solution \cite{luo2018end}.
Its network consists of convolution and LSTM networks, taking raw frames as states and outputting camera actions.
The task is formulated as a Markov Decision Process (MDP) problem, and A3C \cite{mnih2016asynchronous} is adopted as its RL algorithm. 
The experiments are conducted within virtual 3D environments.
The results show that the end-to-end method performs favorably against many traditional trackers with a hand-engineered camera-control module, and also has good generalization capability and potential to transfer to real-world scenarios.
However, this end-to-end method assumes that the target motion pattern is fixed, \emph{i.e.,} the target only moves along a fixed trajectory, which limits the generalization capability of the tracker.

Still based on the single-target single-camera system, the follow-up work addresses the above problem by adding an adversarial mechanism \cite{zhong2018ad}.
Both the tracker and the target are given a partial zero-sum reward and learn to have a better tracking policy or avoid the tracker.
Such a mechanism diversifies the visual appearance and tries to detect the weakness of the tracker, which vice versa yields a more robust and efficient tracker.
Active tracking in a more complex scene with multiple distractors is also studied \cite{activexi}, and attention modules are used to encode the template feature and the historical feature embeddings.

\subsubsection{Multi-camera AOT via deep RL} The multi-camera collaboration system is also investigated.
A pose-assisted method is proposed \cite{li2020pose}, in which cameras share their pose information when tracking a single target to overcome imperfect observations like occlusion.
The system has two controllers named vision-based controller and pose-based controller.
A switcher is used to evaluate the observation and choose which controller to use.
The empirical results in 3D virtual environments show that this method has the capability to deal with complex scenes compared with traditional tracking algorithms.

\begin{figure*}
    \centering
    \includegraphics[width=\columnwidth]{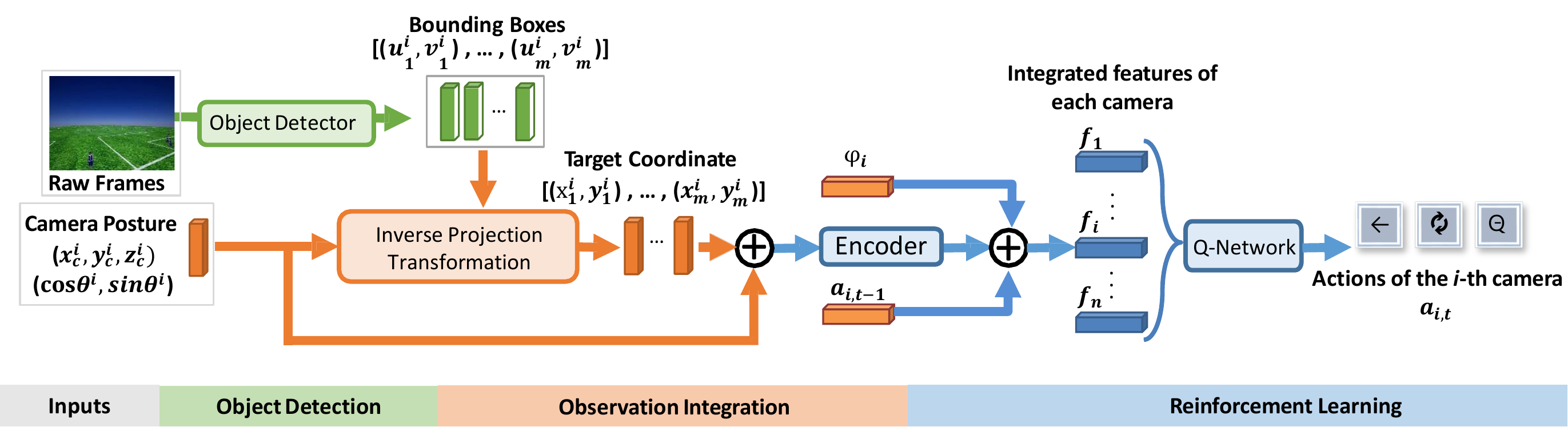}
    \caption{An overview of our multi-agent active tracking system. $\bigoplus$ denotes concatenation and $\varphi_i$ is the one hot code of the $i$-th camera. In each step, each agent gets primitive frame pixels from the environment. These observations are further integrated into joint observation via object detector and inverse projection transformation. Then the Q-Network picks an optimal action according to the current observation and policy. After conducting actions, the environment evolves into the next step, and the rewards generated will be used to update the Q-Network. }
    \label{fig:overview}
\end{figure*}

Despite the progress achieved by previous works in active tracking, the active multi-object tracking (AMOT) task has not been explored yet. 
In this work, this task is formulated as a coverage problem.
A number of cameras are supposed to cover the maximum number of random-walking targets within their view range.
Previous work applies the hierarchical reinforcement learning method on a 2D version of the coverage problem \cite{xu2020learning}, suggesting that multi-agent RL is a potential solution to AMOT. 
However, in the 2D environment, targets are simplified as points, omitting any visual appearances.
Moreover, cameras are assumed to have direct access to the targets' location, which is unrealistic.
Different from that, a 3D multi-object multi-camera environment could be more complicated with raw frames as inputs and imperfect observations like occlusion.

Inspired by previous work, we first build a 3D virtual environment to simulate real-world scenarios and study the AMOT problem on it.
A centralized multi-agent RL algorithm is deployed to learn a proper policy for each camera to take action.
In contrast, our system efficiently deals with AMOT problem given raw frames and has much higher coverage compared with the equal number of fixed cameras.

\section{Method}
In this section, we introduce our system for the AMOT task in detail. We first discuss the problem formulation and then show the whole procedure sequentially in three parts: object detection, observation integration, and reinforcement learning. Figure \ref{fig:overview} illustrates an overview of our method.

\subsection{Problem Formulation}
The active tracking problem mentioned above is formulated as a partially observable multi-agent Markov decision process (POMDP) \cite{KAELBLING199899}.
Traditional POMDP is defined with a 7-tuple $\left\langle S,A,T,R,\Omega,O,\gamma\right\rangle $, which denotes a set of states, actions, conditional transition probabilities, reward function, observations, conditional observation probabilities, and discount factor, respectively.
Specially in our multi-agent system, we denote $I= \left\{I_{1},I_{2},\cdots,I_{n}\right\}$ as the set of agents, each of which corresponds to a camera.
All cameras share the same observation space and action space.
The action space is discrete and described as a vector with 3 dimensions: $a = [a_m, a_r, a_z], a \in A$, denoting action on translation, rotation and zoom, respectively.
Based on the above definition, our multi-agent system is governed with 8 elements, \emph{i.e.,} $\left\langle S,A,T,R,\Omega,O,\gamma,I\right\rangle$.

In step $t$, the $i$-th agent firstly gets its observation $o_{i,t}$ with probability $O\left(s_{i,t}\right)$, where $o_{i,t}\in\Omega$, and $s_{t}= \left\langle s_{1,t},s_{2,t},\cdots,s_{n,t}\right\rangle \in S$.
Then the $i$-th agent sends its observation to the Q-network, and all agents' observations will be further combined into a joint observation $o_{t}=\left\langle o_{1,t},o_{2,t},\cdots,o_{n,t}\right\rangle$.
Concatenated with one-hot code to identify each agent, $o_{t}$ is sent to the Q-network to obtain the estimated Q-value $q_{t,s,a}$ of each action.
And the $i$-th agent's action $a_{i,t}$ is then picked via an epsilon-greedy policy based on the Q-values.
All agents' picked actions compose the joint action $a_{t}=\left\langle a_{1,t},a_{2,t},\cdots,a_{n,t}\right\rangle$ .
Finally, the updated state $s_{t+1}$ will be drawn with probability $T\left(s_{t+1}|s_{t},a_{t}\right)$, and the $i$-th agent will receive a reward $R_{i,t}$.
The ultimate goal for our AMOT problem is to maximize the expected total reward within the whole episode:
\begin{equation}
    \mathbb{E}\left[\sum_{i=1}^n\sum_{t=1}^{T}R_{i,t}\right],
\end{equation}
where $T$ is the number of steps in an episode.

\subsection{Agent State Representation}
The state representation of agents has significant impact on efficiency of reinforcement learning.
In our task, the information originally received from cameras only contains images with raw pixels and locations of each camera, which is hard to learn an effective RL policy. 
To address it, target detection and location alignment are applied sequentially to convert primitive observations to a more compact and reasonable state representation.

\subsubsection{Target Detection}
At the beginning of each step, each camera takes shots of the virtual environment based on current environment state.
To estimate the location of each target, object detection algorithms are adopted to predict bounding boxes based on these raw frames, which act as a part of the observation defined above.
Besides, due to the real-time efficiency of our environment, the object detector is also supposed to output results with high quality and efficiency.

To this end, we use the YoloV4-Tiny model as our object detection model, which is derived from state-of-the-art object detection algorithm YoloV4 with a simplified network structure and reduced parameters \cite{bochkovskiy2020yolov4,wang2021scaledyolov4}.
Compared to YoloV4 which has more than 60 million parameters, the YoloV4-Tiny model contains only 6 million parameters.
This greatly increases the feasibility of deploying this object detector in our active tracking problem.
Our experimental results show that the performance of taking YoloV4-Tiny as an object detector is comparable with that of using ground truth bounding boxes generated from the environment.

\subsubsection{Location Alignment by Inverse Projection Transformation}
However in AMOT task, bounding boxes obtained from object detection merely demonstrate the relative position between the camera and the object but not the absolute position of the object in the 3D environment, lowering the training efficiency of reinforcement learning.
Compared to the bounding boxes, the absolute coordinate of each target in the 3D environment is a better state representation which provides the absolute positions of objects.
Therefore, bounding boxes are mapped to targets' coordinates after detection via inverse projection transformation.
Inspired by the camera calibration technology \cite{786974}, we design inverse projection transformation as an effective solution to convert bounding boxes into coordinates.

According to the pinhole camera model, the pixel-level 2D point position $\left(u,v\right)$ in the frame and 3D point position $\left(x,y,z\right)$ in world coordinates are correlated with the projection equation:
\begin{equation}
    Z_{c}\begin{bmatrix}
        u \\
        v \\
        1 
    \end{bmatrix}
    =\begin{bmatrix}
        f_{x} & 0 & u_{0} & 0 \\
        0 & f_{y} & v_{0} & 0\\
        0 & 0 & 1 & 0
    \end{bmatrix}
    \begin{bmatrix}
        R_{3\times3} & T_{3\times1} \\
        0_{1\times3} & 1
    \end{bmatrix}
    \begin{bmatrix}
        x \\
        y \\
        z \\
        1
    \end{bmatrix},
    \label{con:projection}
\end{equation}
where $R$ and $T$ are the extrinsic parameters of the camera in the world coordinate system.
Since the targets randomly walk on the ground plane, their $z$ coordinates are always zero since the z-axis is set perpendicular to the ground.
In this context, Equation \eqref{con:projection} is simplified into:
\begin{equation}
    Z_{c}\begin{bmatrix}
        u \\
        v \\
        1 
    \end{bmatrix}
    =\begin{bmatrix}
        f_{x} & 0 & u_{0} \\
        0 & f_{y} & v_{0} \\
        0 & 0 & 1 
    \end{bmatrix}
    \begin{bmatrix}
        r_{1} & r_{2} & T_{3\times1} \\
    \end{bmatrix}
    \begin{bmatrix}
        x \\
        y \\
        1 \\
    \end{bmatrix},
\end{equation}
where $r_{1}$ and $r_{2}$ are the first and second columns of matrix $R$.
The intrinsic matrix $K$ is defined as below:
\begin{equation}
    K = \begin{bmatrix}
        f_{x} & 0 & u_{0} \\
        0 & f_{y} & v_{0} \\
        0 & 0 & 1 
    \end{bmatrix}.
\end{equation}
Parameters in the intrinsic matrix contain focal length and principal point, which only depend on the camera's inherent structure, invariant to the camera's position and direction.

In our virtual environment, the origin of the world coordinate system is placed at the center of the ground. 
The camera's intrinsic matrix, pose, and location at each step are known.
Given 2D point position $\left(u,v\right)$ in the frame, its position in world coordinate $x$ and $y$ is calculated by:
\begin{equation}
    \begin{bmatrix}
        x \\
        y \\
        1 \\
    \end{bmatrix}
    =Z_{c}\begin{bmatrix}
        r_{1} & r_{2} & T_{3\times1} \\
    \end{bmatrix}^{-1}
    K^{-1}
    \begin{bmatrix}
        u \\
        v \\
        1 
    \end{bmatrix}.
\end{equation}

\subsubsection{Agent Observation Integration}
The middle point of the bottom margin of the bounding box is used as the 2D point position $\left(u,v\right)$ to estimate the target's world coordinate in our method.
With coordinates obtained, the observation of the $i$-th agent $o_{i}$ is defined as a joint vector with coordinates of detected targets $\left\{ \left\langle p_{1}, q_{1}\right\rangle, \cdots, \left\langle p_{m}, q_{m}\right\rangle \right\}$ and camera's posture information including location $\left\langle x_{i}, y_{i}\right\rangle$, rotation $\alpha_{i}$, zoom scale $z_{i}$, and $L_1$ distance to others $l_{i,j}$.

\subsection{Reward Definition} 
To help our agents learn a competent and robust policy, it's nontrivial to design a proper reward.
Our reward consists of a team reward and an individual reward. 
For the $i$-th agent, the reward $R_{i,t}$ is defined as a weighted sum of the team reward $R_{t}^{T}$ and the individual reward $R_{i,t}^{I}$: 
\begin{equation}
    R_{i,t} = w_{T}\cdot R_{t}^{T} + (1-w_{T})\cdot R_{i,t}^{I} \label{con:reward}.
\end{equation}

\subsubsection{Team Reward}
With $m$ targets in our environment and coverage rate as evaluation, the team reward is defined same as the coverage rate:
\begin{equation}
    R_{t}^{T} = \frac{\sum_{j}\max_{i}v_{i,j,t}}{m},
    \label{eq:teamreward}
\end{equation}
where $v_{i,j,t}$ is the visible flag.
If the ground truth bounding box size of target $j$ in camera $i$, noted as $S_{i,j,t}$, is larger than the whole frame size by a threshold $\mu_{min}$ at step $t$, the value of $v_{i,j,t}$ is set to $1$, otherwise $0$.

\subsubsection{Individual Reward}
The individual reward is the weighted sum of four reward terms:
\begin{equation}
    R_{i,t}^{I} = R_{b,t} + \lambda_{v}R_{v,t} + \lambda_{d}R_{d,t} + \lambda_{p}R_{p,t},
\end{equation}
where $R_{b,t}, R_{v,t}, R_{d,t}$, and $R_{p,t}$ are four terms considering bounding box size, visibility, direction, and position, respectively. In the following, we elaborate each term of the individual reward separately. 

Firstly, the target bounding box is expected to have enough size for tracking. 
Therefore with $S_{i,j,t}$ for the bounding box size of target $j$ in camera $i$ at step $t$ and $S_{0}$ for the frame size, the bounding box reward is also defined:
\begin{equation}
    R_{i,t}^{b} = \min\left\{ 100\sum_{j}\frac{S_{i,j,t}}{S_{0}},\mu_{max} \right\}.
\end{equation}

Secondly, cameras are supposed to cover the maximum number of targets, which coincides with the average coverage as an evaluation metric.
Meanwhile, it's possible that a single target is simultaneously observed by more than one camera, which has no benefits but wastes camera resources.
Considering the above two issues, visibility reward is defined as below:
\begin{equation}
    R_{i,t}^{v} = \sum_{i}\frac{v_{i,j,t}}{\sum_{k}v_{i,j,t}},
\end{equation}
where $v_{i,j,t}$ takes the same definition as that in Eq.~\eqref{eq:teamreward}.

Thirdly, a camera already capturing at least one target should hold the target in its visual field instead of losing it in the near future.
Therefore, the camera is supposed to adjust its pose to track the target in the center of its view, otherwise the target can easily walk out of the visual field.
Thus a direction reward is designed as below:
\begin{equation}
    R_{i,t}^{d} = 1 - \frac{\left|\varDelta\alpha_{t}\right|}{\alpha_{max}},
\end{equation}
where $\varDelta\alpha$ is the angle error between camera orientation and target
direction in yaw angles.
Specifically, if there are more than one target in the frame, the average position is used to represent the target position.

Fourthly, considering a heuristic view that staying in the same position will limit cameras' global perspective, a position reward is designed to keep cameras in distance:
\begin{equation}
    R_{i,t}^{p} = -\max\left\{\frac{d_{max}-\min_{j,j\neq i}d_{i,j,t}}{d_{max}},0\right\},
\end{equation}
where $d_{i,j,t}$ is the position difference between camera $i$ and camera $j$ at step $t$.

\subsection{Centralized Reinforcement Learning} \label{sec:RL}
With state representation and reward function defined above, a centralized value-based multi-agent RL algorithm is used to train our agents, given both estimated target coordinates and camera pose as features. In the following, we discuss the network architecture designed in our algorithm as well as the training strategy. 

\begin{figure}
    \centering
    \includegraphics[width=\columnwidth]{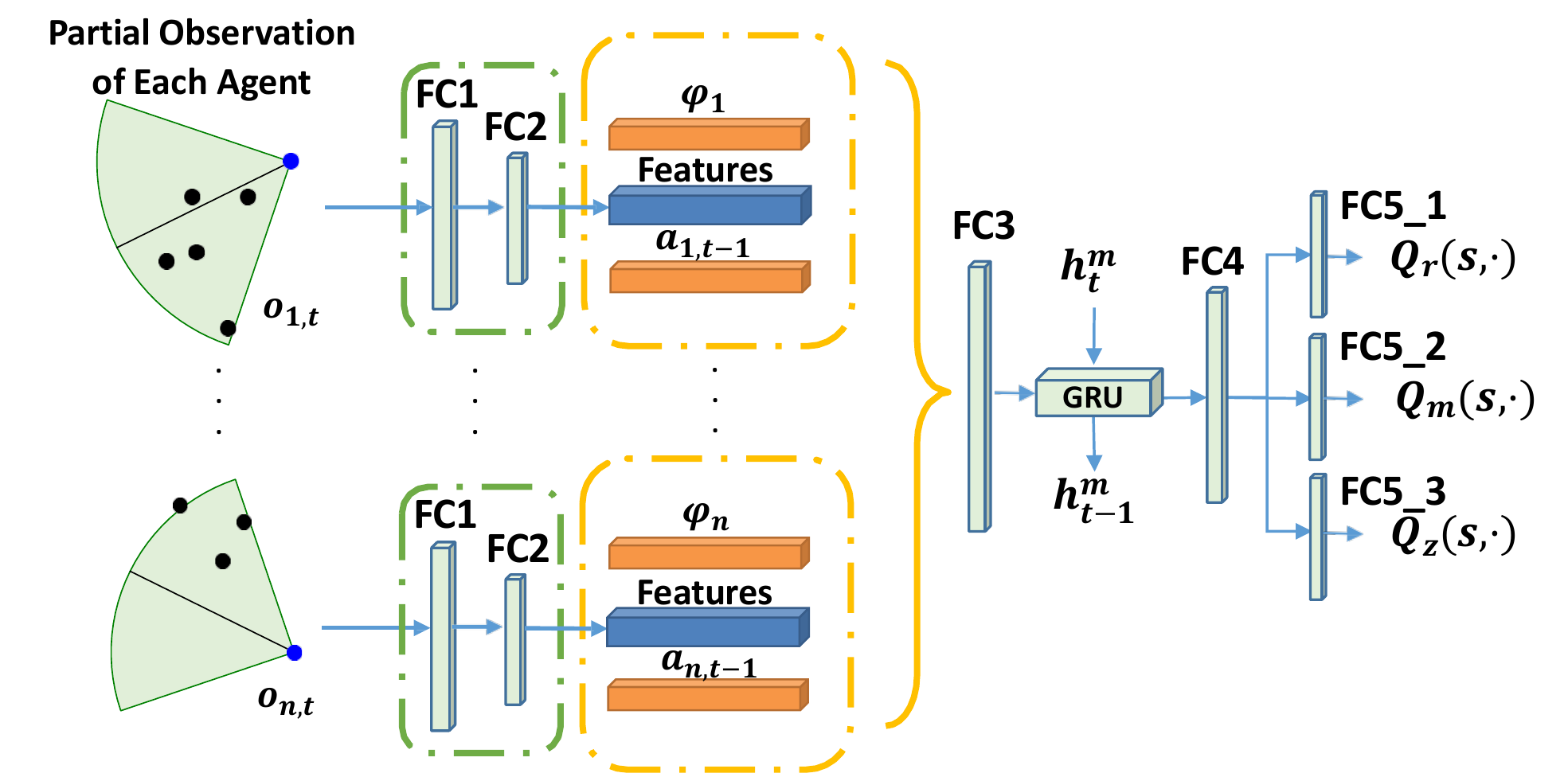}
    \caption{The network architecture of our Q-Network. Note that the FC and GRU represent the Fully Connected layer and the Gated Recurrent Unit, respectively. $\varphi_i$ is the one-hot code used to identify different agents. The i-th camera's output features from the first two FC layers are combined with its one-hot code and last action $a_{t-1}$ and then concatenated with other cameras' features. The joint feature is the input of the FC3, and finally, the network outputs three Q-values of the i-th camera evaluating actions of translation, rotation, and zoom.}
    \label{fig:RLnetwrok}
\end{figure}
\subsubsection{Network Architecture}
In our method, a deep Q-network is adopted to approximate the Q-value function.
All agents share the same Q-network and choose their actions based on the estimated Q-values and epsilon-greedy algorithm.
Figure \ref{fig:RLnetwrok} shows our network architecture.
When obtaining the estimated Q-values of a single agent, partial observations of all agents will be input into the network since these observations are shared.
Each agent's partial observation will be firstly encoded through a unit consisting of two fully connected layers.
The encoded feature is then concatenated with the one-hot code and agent's last action, which enables the network to identify specific agent to estimate Q-values and provides temporal information.
The combined features of all agents are then concatenated together and input into the follow-up layers. 
Finally, the network outputs the Q-values of each possible action of the agent.
Specifically, the last fully connected layer is divided into three branches to accelerate convergence, considering that our action is denoted as a 3-dimensional vector.

\subsubsection{Training Strategy}
Previous works have proven that the conventional deep Q-network algorithm suffers from substantial overestimation \cite{hasselt2010double}.
Thus double Q-learning method \cite{van2016deep} is adopted to update our Q-network.
In double Q-learning, two Q functions: $Q^{A}$ and $Q^{B}$ are stored.
These two functions both provide Q-values of each action given situation $s$.
However, function $Q^{A}$ is updated with the value from the $Q^{B}$ to avoid overestimation:
\begin{equation}
    Q^{A}_{t+1}(s,a) = Q^{A}_{t}(s,a) + \alpha(R_t + \gamma Q^{B}(\hat{s},a^{*})), 
\end{equation}
where $s$ and $\hat{s}$ are the current and next situation, $a$ is the chosen optimal action, and $a^{*}$ is defined as follows: 
\begin{equation}
    a^{*} = \arg\max_{a} Q^{A}(\hat{s},a), 
\end{equation}
In practice, we use two deep Q-networks to learn these two functions.
The networks are updated by the temporal difference error (TD error) $e^{td}$ calculated as follows:
\begin{equation}
    e^{td}_{t} = Q^{A}(s,a) - \left(R_{t} + \gamma Q^{B}(\hat{s},a^{*})\right).
\end{equation}
Then the optimizer uses squared TD error as $L_2$ loss to update our estimate network $Q^{A}$, and parameters of $Q^{A}$ are copied to $Q^{B}$ every $k$ episodes.

\begin{figure*}
    \centering
    \includegraphics[width=\columnwidth]{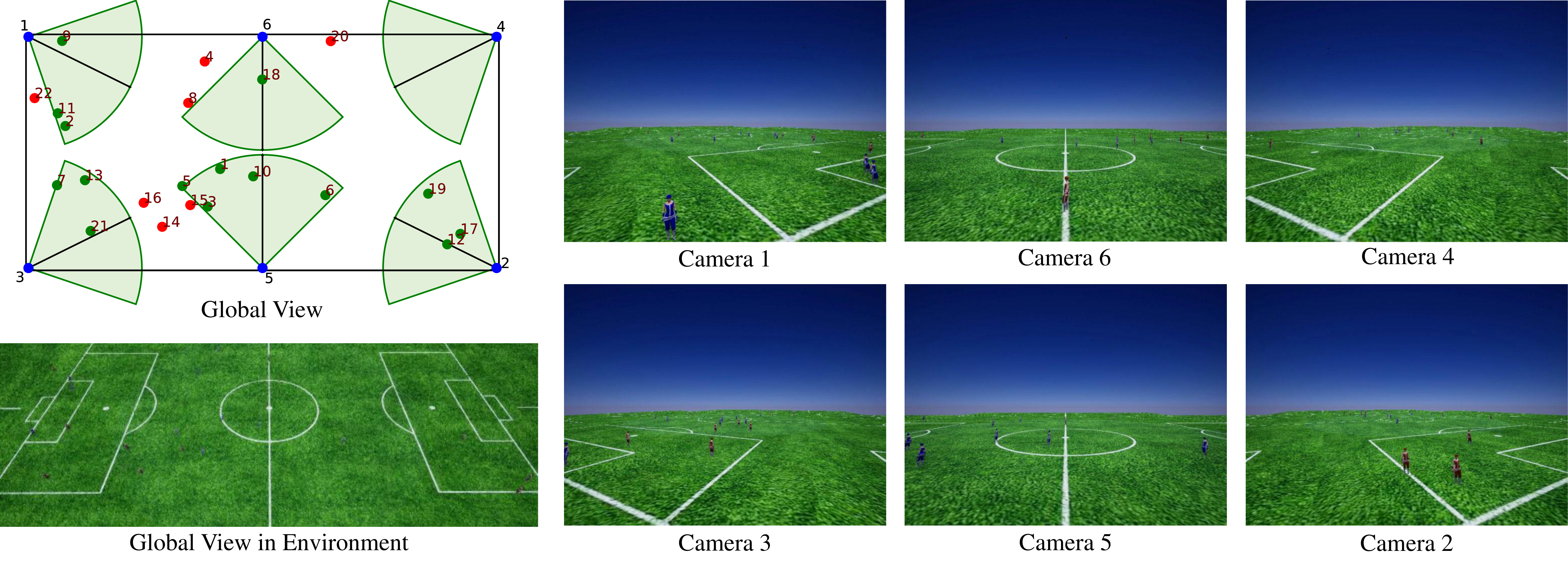}
    \caption{Our 3D environment. The left two figures show the global view while the six figures in the right respectively show the raw observation of six cameras. The size of the ground truth bounding box is used to judge whether an object is covered. And the green points denote covered objects while the red points denote the uncovered ones in the global view. }
    \label{fig:env_and_cam}
\end{figure*}

\section{Experiments}
In this section, we first discuss in detail our environment and settings. Then we evaluate our approach for AMOT task in this environment.

\subsection{Simulation Environment}
Training agents for multi-agent active tracking in real-world scenes is a tough work due to several reasons.
Firstly, it's hard to determine the ground truth to compute reward function or evaluate performance.
Secondly, building a new data set specially for active tracking is also expensive due to numerous possible states.
Thirdly, our agent needs to interact with the environment frequently to optimize its policy, while it may suffer from high-cost trial-and-error.

To avoid these problems, a virtual environment is built which simulates real-world multi-agent scenes to train our agents.
The environment and our code are released to the public for convenience of repeating our results and conducting further study~\footnote{\url{https://drive.google.com/drive/folders/1K4PEIohiZAyJ123lBESJO1D8lDy977in?usp=sharing}}. 
And to make the training realistic, the simulation operates in real-time.
The capability of generalizing trackers trained in the virtual environment to real-world scenes has already been shown in previous works \cite{zhong2018ad,luo2018end}.
The 3D virtual environment is built on the Unreal Engine, with UnrealCV\cite{qiu2017unrealcv,gymunrealcv2017} and OpenAI Gym\cite{1606.01540} which provides convenient APIs for interaction between the algorithm and the environment.

\begin{figure}
    \centering
    \includegraphics[width=\columnwidth]{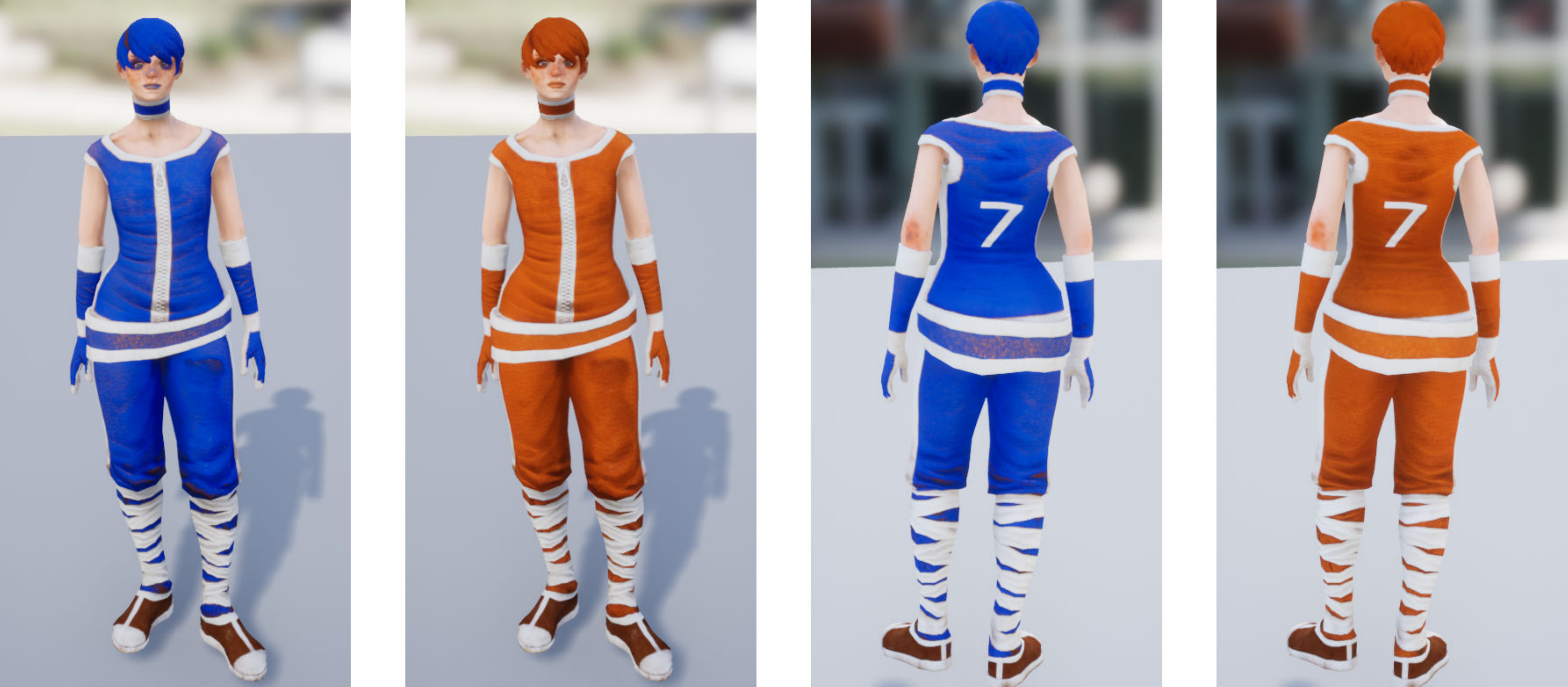}
    \caption{Two different appearances identify the two teams in our environment. The number in the back varies among different targets.}
    \label{fig:character}
\end{figure}

Our environment simulates a real-world soccer court with twenty-two human-like targets and six mobile cameras, shown in Figure~\ref{fig:env_and_cam}.
The size of the scope of target activity is $10000 \times 5000$ world units, where 100 world units refer to 1 meter in the real world.
The targets in our environment are divided into two teams, and the appearances within the same team have little difference.
In our environment, each target is a human-like character with the appearance shown in Figure~\ref{fig:character}.
To mimic real-world soccer match scenes, targets are divided into two teams with different colors, while targets in the same team have no difference except the number on the back.
At the beginning of each episode, each target is randomly placed within the field.
To implement random walking, we randomly set a reachable point in the field as a destination for each target.
Then the target walks at a constant speed towards that destination.
If the destination is reached or more than 15 seconds are costed, another point will be picked as the destination for that target.

The action space of each camera is combined with three independent dimensions.
Agent respectively chooses actions from translation (move clockwise, keep still, move anticlockwise), rotation (turn left, keep still, turn right) and scale (zoom in, keep still, zoom out).
Then the three separate actions are combined into joint action.
Therefore, each agent holds in total 27 different actions.
The translation action moves the camera clockwise or anticlockwise along the border by 100 world units.
The rotation action manipulates the direction of the camera by 10 degrees. 
And the zoom action changes the zoom scale of the camera by 10\%.

At the beginning, the six cameras are evenly placed on the border of the court, with a height of 500 world units. 
The cameras will catch a frame with a resolution of $640 \times 480$, and its view angle is set to 90 degrees.
Our environment also allows our cameras to observe in object mask method, so it's easy to get the ground truth of targets' bounding box.
Therefore, we use the bounding box size as the criteria of whether an object is covered in the visual field of a camera, which is more reasonable than the relative distance between the target and the camera.

\begin{figure*}
    \centering
    \includegraphics[width=\columnwidth]{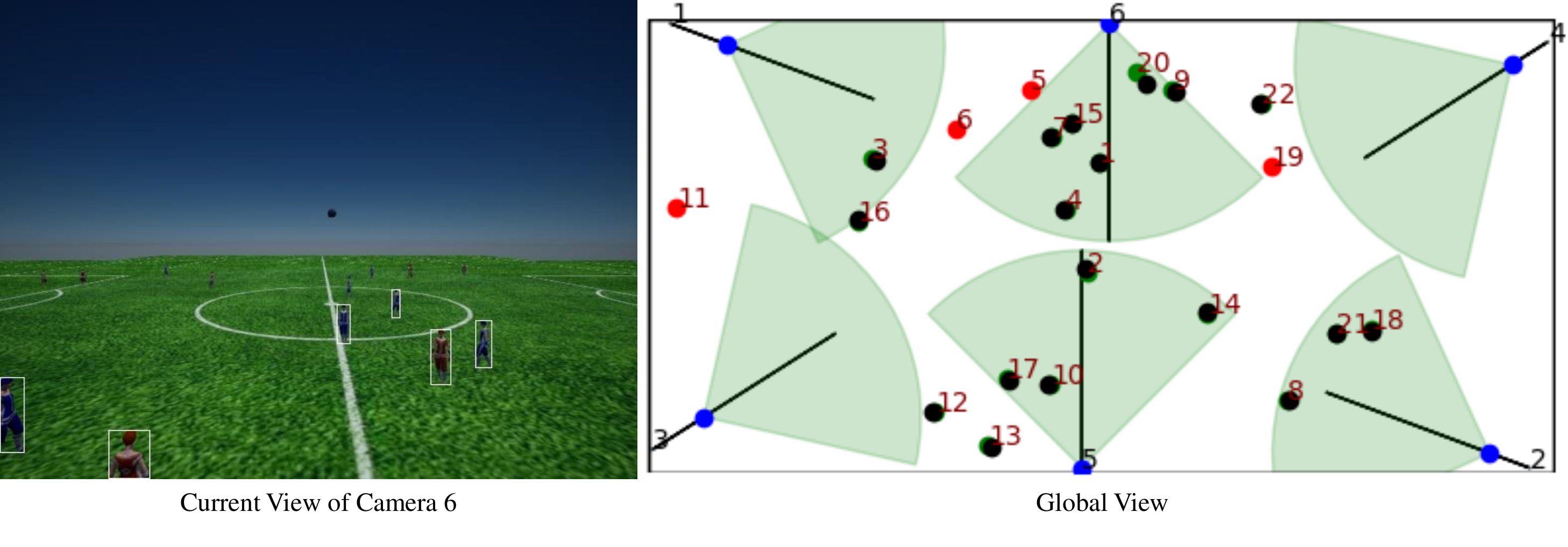}
    \caption{An example of inverse project transformation. The black point in the right graph refers to the estimated location of each observed target while the green point under that denotes the ground truth location.}
    \label{fig:ipt}
\end{figure*}

\subsection{Settings}
\subsubsection{Evaluation Metric}
In our environment, with limited numbers of cameras, the joint visual field is unable to cover the whole area.
So if the cameras are fixed, in most cases there are always some targets out of view.
Therefore the ultimate goal is to train these cameras to cover as many targets as possible.
For this purpose, the average coverage rate is used for the metric:
\begin{equation}
\gamma = \frac{1}{m}\sum_{j=1}^{m}\frac{1}{T}\sum_{t=1}^{T}v_{i,j,t},
\end{equation}
where $v_{i,j,t}$ is define in Equation \eqref{eq:teamreward}, $\gamma$ is the average coverage rate, $m$ and $T$ refer to number of targets and episode length, respectively.
And the bounding box area is used as a threshold to determine whether a target is captured by a camera.
Existing metrics in passive multi-object tracking (MOT) or single-object active tracking tasks seem plausible for the AMOT task but are actually not applicable. 
For example, the Multiple Object Tracking Accuracy (MOTA) is a fair metric in MOT. 
It is defined by number of misses, false positives, and mismatches.
However, it can only evaluate the effectiveness of the detector. 
Camera control policy have little influence on MOTA since it don't change the performance of detectors but their poses.
Therefore, we only use coverage rate as the evaluation metric. 

\begin{table}[!t]
\caption{The details of parameters of the rewards.\label{table1}}
\centering
\resizebox{0.5\textwidth}{!}{
\begin{tabular}{lll}
\hline
 Parameters & Physical Meaning & Value \\
 \hline
 $\omega_{T} $ & Team reward weight & 0.4 \\
 $\alpha_{max} $ & Max angle in direction reward & $\pi$/4 \\
 $ \mu_{max}$ & Max size proportion in bounding box reward & 0.2 \\
 $ \mu_{min}$ & Min size proportion in bounding box to be counted observed & 0.0005\\
 $d_{max}$ & Max camera distance in position reward & 5000 \\
 $\lambda_{v} $ & Vision reward weight & 0.8 \\
 $\lambda_{d}$ & Direction reward weight & 0.2 \\
 $\lambda_{p}$ & Position reward weight & 0.2 \\
 \hline
\end{tabular}
}
\end{table}

\subsubsection{Baseline}
Due to differences in environment setting, task goal, and problem formulation, simply transferring existing algorithms in other tasks like coverage problem in sensor deployment \cite{sensor} is not fair and thus unconvincing. 
Besides, previous work in active tracking only involves single-target tracking tasks, with no  datasets or algorithms for AMOT.
Thus we provide fixed cameras as baselines for our mobile cameras to compare with.
These fixed cameras are evenly distributed at the border and oriented towards the center point of the court.

\subsubsection{Hyper Parameters}
For our RL algorithm, the learning rate is 0.0005. 
The reward discount factor $\gamma$ is 0.99. 
The batch size which is the number of episodes to train on is set to 32.
In the epsilon greedy policy, factor $epsilon$ starts from 1.0 and finished at 0.1 with an annealing time of 50000 steps.
One episode lasts for 100 steps, and after the final step the environment will be reset, \emph{i.e.,} the position of each target will be randomly set, and the position and pose of each camera will be initialized.
Since we use double Q-learning, the target network $Q^{B}$ updates every 100 episodes.
The episode length is set to 100 steps.
For evaluation, we train each model for 500k steps, \emph{i.e.,} 5k episodes.
The parameters of the rewards are present in Table \ref{table1}.

\begin{figure*}
    \centering
    \includegraphics[width=\columnwidth]{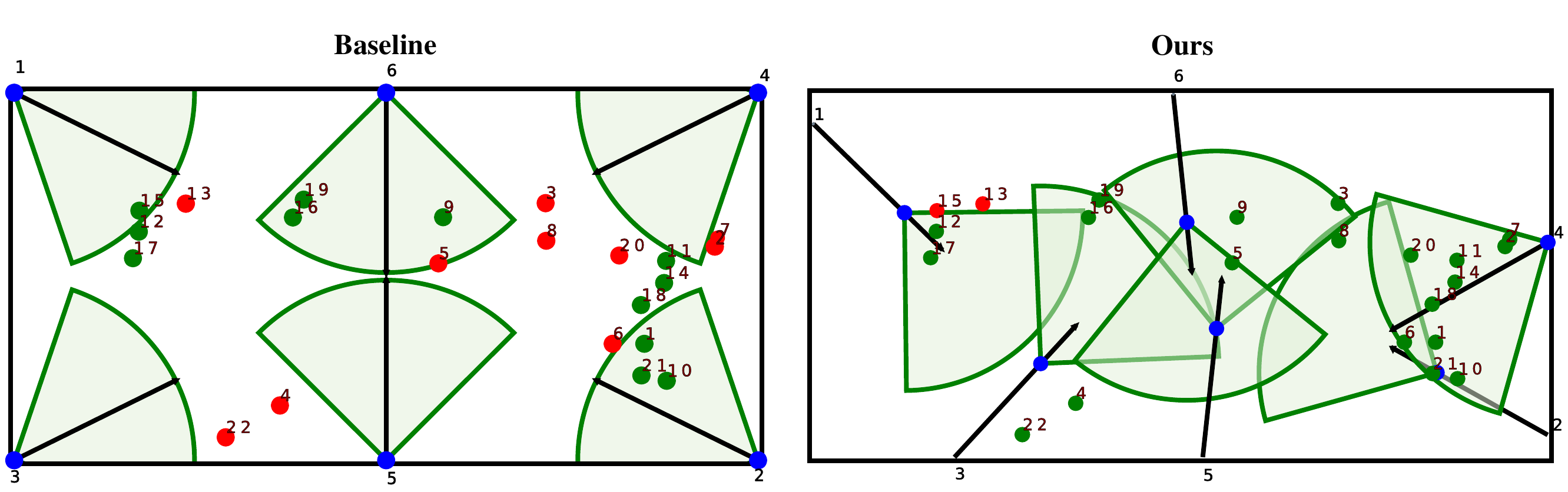}
    \caption{Global view of the environment when applying baseline and our method. For comparison, targets are placed at the same location. Since the bounding box size instead of distance is the only criterion of coverage, the green wedge in this figure is only an approximate visual field of each camera. So targets 4 and 22 are green in the right graph though out of any wedge.}
    \label{fig:example}
\end{figure*}
\subsection{Performance Evaluation and Discussion}
We quantitatively evaluate the performance of our method and compare it to the baseline method.
To show the necessity of inverse projection transformation, we also compare our method with the method that directly uses bounding boxes as Q-network inputs.
Ablation analysis is also conducted to show the effectiveness of the proposed different individual rewards.
\subsubsection{Evaluation}
\begin{table}[!t]
\caption{Comparative results in ``soccer court'' environment. Note that CR represents Coverage rate,  ``Ours+'' refers to our method with ground truth bounding boxes replacing object detector, and ``Ours-'' refers to our method without inverse projection transformation. \label{table2}}
\centering
\setlength{\tabcolsep}{0.1\columnwidth}{
\begin{tabular}{ll}
\hline
 Method & CR(\%) \\
\hline
Baseline & $63.0\pm4.5$ \\
Ours & $71.9\pm5.8$\\
Ours+ & $72.1\pm5.0$\\
Ours- & $66.9\pm5.8$\\
\hline
\end{tabular}
}
\end{table}
Our agents are trained and tested with the settings mentioned above.
We compare our system with baseline and other modified methods shown in Table \ref{table2}. To compare the effectiveness of the object detector, we compare our method with ``Ours+", in which precise bounding boxes of each target are directly provided by the environment instead of the detector. 
Moreover, to demonstrate the effectiveness of inverse projection transformation for coordinate alignment, we also compare our method with the two-stage tracking method ``Ours-".
In ``Ours-", the step of inverse projection transformation is omitted, thus the bounding boxes detected in each camera are the inputs of the Q-networks, instead of coordinates.
Considering the uncertainty brought by initialization, we conduct 100 runs in each method and report the mean and standard deviation of the coverage rate.

The results demonstrate that our agents have the capability to move and track actively.
Thus, when targets are nearly out of view, our agent can take action to approach the target, trying to keep it in its capture.
Meanwhile, the fixed cameras are unable to deal with these situations and keep losing the target.
The results also show that using object detection and ground truth bounding boxes has little difference in performance, which proves the effectiveness of using Yolov4-tiny as our object detector.
It also demonstrates that using inverse projection transformation for alignment notably improves the convergence speed and also achieves better performance.
Figure~\ref{fig:example} illustrates an example of the difference between ours and the baseline method.
The fixed cameras have plenty of blind areas where targets can never be captured.
Meanwhile, our agents track objects actively when they are about to move out of their visual field, and cooperate to cover maximum numbers of targets.

To further verify the correctness of the Inverse Project Transformation technique, we test for 1000 steps to compare the calculated coordinates with the ground truth.
The results show that the average Euclidean distance between them is 29.6 world units with an std. of 8.04.
Considering that the size of the scope of target activity in our environment is $10000 \times 5000$ world units, the estimated error is negligible.
The main cause of the error is the incomplete bounding box of the partially observed target.
Figure~\ref{fig:ipt} shows an example of inverse project transformation.
For most of the targets, their estimated locations are close to the ground truth, as the black points nearly cover the green points.
However, for target 20, the estimation is not perfect since the object is partially captured by camera 6.
The real position of it is closer to the camera.

\begin{table}[!t]
\caption{Results of ablation study. Note that CR represents Coverage rate. We adjust the reward structure for each compared method. Models are trained for 200k steps.\label{table:ablation}}
\centering
\setlength{\tabcolsep}{0.02\columnwidth}{
\begin{tabular}{lllllll}
\hline
 Method & $R^T$ & $R^v$ & $R^d$ & $R^b$ & $R^p$ & CR(\%) \\
\hline
Ours & $\surd$ & $\surd$ & $\surd$ & $\surd$ & $\surd$ & $63.6\pm5.4$\\
Ours - Vis & $\surd$ &  & $\surd$ & $\surd$ & $\surd$ & $61.5\pm5.3$\\
Ours - Dir & $\surd$ & $\surd$ &  & $\surd$ & $\surd$ & $55.5\pm5.3$\\
Ours - Box & $\surd$ & $\surd$ & $\surd$ &  & $\surd$ & $57.8\pm5.4$\\
Ours - Pos & $\surd$ & $\surd$ & $\surd$ & $\surd$ &  & $61.0\pm6.9$\\
Decentralized &  & $\surd$ & $\surd$ & $\surd$ & $\surd$ & $61.6\pm 5.5$\\
Team-only & $\surd$ &  &  &  &  & $55.7\pm 4.9$\\
\hline
\end{tabular}
}
\end{table}
\subsubsection{Ablation Analysis}
Our final reward is a weighted sum of individual rewards and the team reward.
In Section \ref{sec:RL} multiple individual rewards are introduced in our reward definition: visibility, direction, bounding box, and position reward.
These four types of rewards are significant for our agents to learn an appropriate collaboration policy.
Therefore we add an ablation analysis to illustrate the effectiveness of these four rewards and combination of the individual rewards and the team reward.
In each method, we remove one of these four rewards from our individual reward.
As shown in Table~\ref{table:ablation}, performance drops considerably when discarding any of these four rewards, or using only team reward or individual reward.
In other words, fusing individual rewards and the team reward is effective, and each type of individual rewards is necessary to help our method achieve a higher coverage rate.

\section{Conclusion}
In this paper, we first formulate the AMOT task as a POMDP problem and then propose the coordinate-aligned multi-camera collaboration method for it.
Moreover, we establish a 3D environment simulating real-world multi-object tracking scenes to train and evaluate our method.
The specifically designed reward and centralized multi-agent reinforcement learning network enable our agent to learn an optimal collaboration policy. 
To address the partial observation integration problem, we leverage inverse projection transformation as an intermediate step, converting bounding boxes into aligned coordinates.
Empirical results show that our method outperforms the traditional method with fixed cameras by achieving a higher coverage rate, and validate the effectiveness of our object detector and inverse projection transformation step.

\bibliographystyle{unsrt}  
\bibliography{references}  





\end{document}